# AN UNSUPERVISED TENSOR-BASED DOMAIN ALIGNMENT

*Chong Hyun Lee, Kibae Lee, and Hyun Hee Yim*

Jeju National University, South Korea

**ABSTRACT**

We propose a tensor-based domain alignment (DA) algorithm designed to align source and target tensors within an invariant subspace through the use of alignment matrices. These matrices along with the subspace undergo iterative optimization of which constraint is on oblique manifold, which offers greater flexibility and adaptability compared to the traditional Stiefel manifold. Moreover, regularization terms defined to preserve the variance of both source and target tensors, ensures robust performance. Our framework is versatile, effectively generalizing existing tensor-based DA methods as special cases. Through extensive experiments, we demonstrate that our approach not only enhances DA conversion speed but also significantly boosts classification accuracy. This positions our method as superior to current state-of-the-art techniques, making it a preferable choice for complex domain adaptation tasks.

## 1. INTRODUCTION

Signal processing–based artificial intelligence systems exhibit excellent performance when the training and testing data are collected under identical conditions. However, in real-world applications, differences in devices or environments often cause discrepancy of data distributions. This domain shift problem makes it difficult for models to generalize to new environments, leading to performance degradation. To address this issue, domain adaptation techniques have been actively studied to reduce the distribution gap between the source and target domains [1], [2], [3].

To tackle this challenge, various methods such as Principal Component Analysis (PCA), Subspace Alignment (SA) [4], and Correlation Alignment (CORAL) [5] have been proposed. However, these approaches vectorize tensor-structured data, thereby failing to fully exploit multi-dimensional structural characteristics [6], [7]. To overcome this limitation, Tensor-Aligned Invariant Subspace Learning (TAISL) [8] and other algorithms [9], [10], [11] have been introduced, which project data into a common invariant subspace to reduce domain discrepancy.

Nevertheless, TAISL imposes orthogonality constraints on the projection matrices (Stiefel manifold), limiting representational flexibility and making it difficult to capture complex domain variations.

In this paper, we propose a tensor-based subspace learning algorithm which can be regarded as extended TAISL [9]. The proposed algorithm incorporates oblique manifold constraints in subspace learning to relax the orthogonal constraint and includes regularization terms for preserving variances of both source and target tensors [12]. This gives representational flexibility and leaning capability by capturing complex characteristics of source and target domain tensor data. TAISL and other algorithms can be regarded as special case of the proposed framework when Stiefel manifold on subspace learning and preservation of source tensor are used. Consequently, this generalization gives faster convergence in domain alignment (DA) and improved classification performance as well. Through experiments on visual and acoustic datasets, we demonstrate the effectiveness of the proposed algorithm. Also, reliability of the proposed algorithm is maintained even in limited target tensor size and also exhibits robustness to label noise. Furthermore, compared to state-of-the-art tensor-based and vector-based domain adaptation methods, it achieves superior classification accuracy.

## 2. DOMAIN ALIGNMENT ALGORITHM

We propose a tensor domain alignment (TDA) algorithm that aligns source and target tensors in a common subspace through Tucker decomposition [7] as TAISL does. Let $N_s$ source samples $\{\mathcal{X}_s^n\}_{n=1,\cdots,N_s}$ be given, where each sample is a $K$-mode tensor $\mathcal{X}_s^n \in \mathbb{R}^{d_1 \times \cdots \times d_K}$. For simplicity, these samples are stacked into a $(K+1)$-mode tensor $\mathcal{X}_s \in \mathbb{R}^{d_1 \times \cdots \times d_K \times N_s}$. Similarly, let $\mathcal{X}_t \in \mathbb{R}^{d_1 \times \cdots \times d_K \times N_t}$ denote the set of $N_t$ target samples. The source and target tensors are projected by matrices $\mathcal{M} = \{M_k\}_{k=1,\cdots,K}$, where each $M_k \in \mathbb{R}^{d_k \times d_k}$, and aligned within a common subspace defined by $\mathcal{U} = \{U_k\}_{k=1,\cdots,K}$, where each $U_k \in \mathbb{R}^{d_k \times r_k}$ representing the mode-$k$ subspace of tensor rank $r_k$.

The proposed objective function is formulated as

$$\min_{\mathcal{M},\mathcal{U}} J(\mathcal{M},\mathcal{U}) + \lambda h(\mathcal{M}), \quad \text{s.t. } U_k^{\mathrm{T}} U_k = I, \operatorname{diag}(M_k M_k^{\mathrm{T}}) = 1, \{k=1,\cdots,K\} \tag{1}$$

where $J(\mathcal{M}, \mathcal{U})$ denotes the alignment loss, and $h(\mathcal{M})$ is a regularization term weighted by $\lambda$ to preserve variance. The loss $J(\mathcal{M}, \mathcal{U})$ enforcing alignment of source and target tensors via joint transformation through $\mathcal{M}$ and $\mathcal{U}$, is defined as

$$J(\mathcal{M}, \mathcal{U}) = \|[\![\mathcal{X}_s; \mathcal{M}]\!] - [\![G_s; \mathcal{U}]\!]\|_F^2 + \|[\![\mathcal{X}_t; \mathcal{M}]\!] - [\![G_t; \mathcal{U}]\!]\|_F^2, \quad (2)$$

where, the operator $[\![\cdot; \cdot]\!]$ denotes the Tucker product applied to modes $k = 1, \cdots, K$, excluding the $(K+1)$-th mode and the core tensors $G_s = [\![X_s; \mathcal{U}^T]\!]$ and $G_t = [\![X_t; \mathcal{U}^T]\!]$ represent the source and target projections in the shared subspace. The regularization $h(\mathcal{M})$ is defined as

$$h(\mathcal{M}) = \left\| [\![[\![\mathcal{X}_s; \mathcal{M}]\!] \mathcal{M}^\dagger ]\!] - \mathcal{X}_s \right\|_F^2 + \left\| [\![[\![\mathcal{X}_t; \mathcal{M}]\!] \mathcal{M}^\dagger ]\!] - \mathcal{X}_t \right\|_F^2, \quad (3)$$

where $\mathcal{M}^\dagger = \{M_k^\dagger\}_{k=1,\cdots,K}$ denotes the set of mode-wise Moore-Penrose pseudo-inverses. The $h(\mathcal{M})$ preserves the variance of the original tensors by minimizing the reconstruction error between the input and the recovered tensors via $\mathcal{M}^\dagger$. The pseudo-inverse is used because of the oblique manifold constraint, $\text{diag}(M_k M_k^T) = 1$, which does not guarantee orthogonal transformation but increases the flexible alignment than the Stiefel manifold. It, however, may yield variance dispersion as expressed in (3).

The proposed method can be simplified to the TAISL [8], if the $\text{diag}(M_k M_k^T) = 1$ in (1) is replaced by $M_k M_k^T = I$, the target tensor is not projected in (2), and the second term in (3) is removed. Similarly, if only (1) and (3) are modified with the same manner while keeping (2) unchanged, then it becomes the cost function proposed in [8]. In contrast to prior methods that focus on minimizing target shifts by aligning source data toward the target domain, the proposed approach considers both source and target domains during alignment.

To solve the optimization problem defined in (1), we adopt an alternating minimization strategy. Owing to the interdependence between $\mathcal{M}$ and $\mathcal{U}$, direct joint optimization is computationally intractable. To address this, the objective function is decomposed into two subproblems by alternately fixing one variable set and optimizing the other. This iterative procedure is repeated until convergence.

**Optimize $\mathcal{U}$ given $\mathcal{M}$**: If the projection matrices $\mathcal{M}$s are fixed, then $J(\mathcal{M}, \mathcal{U})$ can be formulated as

$$\min_{\mathcal{U}} J(\mathcal{M}, \mathcal{U}), \quad \text{s.t. } U_k^T U_k = I, \{k = 1, \cdots, K\}, \quad (4)$$

Let $\mathcal{Z}_s = [\![\mathcal{X}_s; \mathcal{M}]\!]$ and $\mathcal{Z}_t = [\![\mathcal{X}_t; \mathcal{M}]\!]$ denote the transformed source and target tensors, respectively. These tensors are concatenated along the $(K+1)$-th mode to form $\mathcal{Z}_{cat} \in \mathbb{R}^{d_1 \times \cdots \times d_K \times (N_s + N_t)}$. The orthogonal subspace matrices $\mathcal{U}$ can then be efficiently obtained by applying the Tucker decomposition using the alternating least square (ALS) algorithm to $\mathcal{Z}_{cat}$ [7].

**Optimize $\mathcal{M}$ given $\mathcal{U}$**: If the projection matrices $\mathcal{U}$s are fixed, then subproblem with respect to $\mathcal{M}$ is formulated as follows:

$$\min_{\mathcal{M}} J(\mathcal{M}, \mathcal{U}) + \lambda h(\mathcal{M}), \quad \text{s.t. } \text{diag}(M_k M_k^T) = 1, \{k = 1, \cdots, K\}, \quad (5)$$

As described in [8], the alignment loss $J(\mathcal{M}, \mathcal{U})$ can be reformulated using the mode-$k$ unfolding matrices of the tensors and it can be written as

$$J_k(M_k, U_k) = \left\| M_k X_{s(k)} M_{\setminus k} - \hat{X}_{s(k)} \right\|_F^2 + \left\| M_k X_{t(k)} M_{\setminus k} - \hat{X}_{t(k)} \right\|_F^2, \quad (6)$$

where $X_{s(k)}$ and $X_{t(k)}$ denote the mode-$k$ unfolding matrices of the source and target tensors, respectively. The matrices $\hat{X}_{s(k)}$ and $\hat{X}_{t(k)}$ correspond to the mode-$k$ unfolding matrices of the reconstructed tensors $\hat{\mathcal{X}}_s = [\![G_s; \mathcal{U}]\!]$ and $\hat{\mathcal{X}}_t = [\![G_t; \mathcal{U}]\!]$, respectively. The term $M_{\setminus k}$ denotes the Kronecker product of all projection matrices except $M_k$. Similarly, the regularization term $h(\mathcal{M})$ can be decomposed with respect to each mode. Following the derivation in [8], the mode-$k$ component is given by

$$h_k(M_k) = -\left(\text{tr}(X_{s(k)}^T M_k^\dagger M_k X_{s(k)}) + \text{tr}(X_{t(k)}^T M_k^\dagger M_k X_{t(k)})\right). \quad (7)$$

Combining the mode-$k$ formulations in (6) and (7), the subproblem with respect to $M_k$ becomes

$$\min_{M_k} J_k(M_k, U_k) + \lambda h_k(M_k), \quad \text{s.t. } \text{diag}(M_k M_k^T) = 1, \{k = 1, \cdots, K\}. \quad (8)$$

Here, we update each $M_k$ by using Riemannian gradient descent algorithm as in [12].

## 3. RESULTS

To evaluate the performance of the proposed algorithm, we conduct experiments on visual and audio datasets obtained different situation. The source and target domain datasets are obtained by dividing image quality (for example, clean and blurred images) and by using different recording devices. After tensor alignment process, a classifier trained solely on the source domain is applied to the target domain to validate the proposed algorithm. Classification accuracy on the target data is used as the primary evaluation metric and the average intra-class distance is computed before and after alignment to assess the improvement in cross-domain geometric consistency.

## 3.1. Dataset and implementation

In experiments using MNIST [13], we consider one source domain and three target domains of *fog* and *motion blur* variants from MNIST-C [14], and MNIST-M [15]. The experiments of MNIST and MNIST-C are conducted on dataset caused by grayscale image degradation, whereas the experiment of MNIST-M on dataset different color space and style. For each experiment, 20 samples per class are randomly selected from both the source and target domains, resulting in tensors $\mathcal{X}_s, \mathcal{X}_t \in \mathbb{R}^{28\times28\times200}$. In the case of MNIST-M, where the images are RGB, the tensor dimensions are $\mathcal{X}_s, \mathcal{X}_t \in \mathbb{R}^{28\times28\times3\times200}$.

For the audio domain dataset, the TUT Urban Acoustic Scenes 2018 Mobile dataset [16] is used. Recordings from devices B and C are assigned as the source and target domains. For training DA, we pair respectively B and C by 1:1 pair and random pair. Each sample is represented as a 128-band log-Mel energies computed using a 40 ms window with no overlap. A total of 30 samples per class are extracted from each domain, yielding tensors $\mathcal{X}_s, \mathcal{X}_t \in \mathbb{R}^{128\times250\times300}$. Fig. 1 illustrates source and target samples selected from the same class using the random pairing strategy.

We use the following baselines: No adaptation (NA), principal component analysis (PCA), subspace alignment (SA) [4], correlation alignment (CORAL) [5], NTSL [8], and TAISL [8]. The extended version of TAISL proposed in [9] is referred to as E-TAISL(S), and its variant with oblique constraints is denoted as E-TAISL(O). The proposed algorithm is evaluated under both the Stiefel and oblique manifold settings, denoted as TDA(S) and TDA(O), respectively.

For TAISL-based methods, the regularization parameter is fixed at $\lambda = 10^{-4}$. The number of learning iterations is set to 80 for both the MNIST-C and MNIST-M experiments with a step size of $10^{-6}$ for Riemannian gradient descent algorithm. For the audio experiments, the iteration count is set to 50 and the step size of $10^{-7}$. The tensor rank $r_k$ for each mode is determined to preserve 99.0% of the total variance by source image tensor and 99.9%. of source audio tensor.

For image experiments, the classifier is implemented as a shallow convolutional neural network (CNN) [17] consisting of a single convolutional layer with 64 filters of size 5×5. In the audio experiments, the same architecture is used except that the convolutional kernel size is increased to 7×7.

## 3.2. Evaluation

Table I reports the average classification accuracy over ten trials on the image tensor. Simple baselines such as NA and PCA achieve relatively low accuracy, while subspace-based methods (SA [4], CORAL [5], NTSL [8], and TAISL [8]) yield improved results and E-TAISL [9] further enhances the performance of TAISL. The proposed TDA achieves the highest accuracy across all target domains and the TDA(O) consistently outperforms other variants, demonstrating the effectiveness of the oblique constraint. Table II presents the results on the audio tensor with which we can observe similar performance obtained with image tensor.

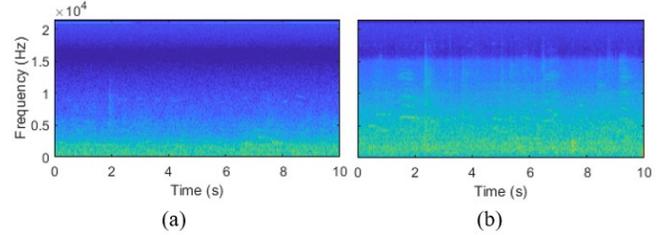

**Fig. 1.** Audio samples from (a) the source domain and (b) the target domain.

TABLE I
AVERAGE ACCURACY (%) ON THE IMAGE TENSOR

| Method | Target data | | |
|---|---|---|---|
| | MNIST-C | | MNIST-M |
| | Fog | Blur | |
| NA | 56.1±2.9 | 64.1±4.1 | 33.2±3.9 |
| PCA | 77.8±2.8 | 65.3±3.5 | 39.1±2.7 |
| SA [4] | 76.8±3.6 | 66.0±1.5 | 40.3±4.2 |
| CORAL [5] | 65.5±4.6 | 67.1±2.9 | 42.0±4.2 |
| NTSL [8] | 75.2±3.2 | 67.3±2.8 | 43.5±3.0 |
| TAISL [8] | 75.5±1.3 | 69.0±3.5 | 44.5±2.6 |
| E-TAISL(S) [9] | 76.9±2.1 | 69.5±1.8 | 46.1±2.3 |
| E-TAISL(O) | 77.0±2.2 | 69.8±3.3 | 47.4±2.0 |
| TDA(S) | 78.9±1.7 | 70.7±3.2 | 50.2±2.4 |
| TDA(O) | 79.2±2.2 | 71.1±1.9 | 53.1±2.4 |

TABLE II
AVERAGE ACCURACY (%) ON THE AUDIO TENSOR

| Method | Target data | |
|---|---|---|
| | 1:1 pair | Random pair |
| NA | 47.3±2.5 | 38.3±3.7 |
| PCA | 59.5±1.7 | 44.0±1.9 |
| SA [4] | 60.3±3.2 | 47.1±2.1 |
| CORAL [5] | 71.3±4.6 | 51.3±3.0 |
| NTSL [8] | 80.8±3.5 | 53.3±3.2 |
| TAISL [8] | 82.1±1.0 | 55.2±1.7 |
| E-TAISL(S) [9] | 82.9±2.9 | 56.7±3.8 |
| E-TAISL(O) | 83.3±2.0 | 60.3±1.9 |
| TDA(S) | 84.5±1.5 | 62.0±1.2 |
| TDA(O) | 85.7±1.0 | 62.7±1.4 |

Fig. 2 presents a t-SNE [18] visualization of the feature distributions for the MNIST-to-MNIST-M experiment, illustrating that TDA(O) exhibits superior domain adaptation performance compared to TAISL. The intra-class distances relatiive to the baseline NA are reduced by 5.5% with TAISL, 8.8% with the proposed TDA(S), and 11.1% with the proposed TDA(O).

Fig. 3 illustrates the loss and classification accuracy over iterations. The results show that TDA(S) converges faster, while TDA(O) exhibits the fastest convergence. This behavior can be attributed to the advantage of the oblique constraint, which enables more effective alignment with fewer iterations. Fig. 4 illustrates the robustness of the proposed methods under varying target-to-source sample ratios in the MNIST-to-MNIST-M setting and audio dataset. As shown in Fig. 4(a) and (b), the classification accuracy of all methods declines as the number of target samples decreases. However, the TDA(O) consistently outperforms the baselines across all settings, demonstrating its robustness under limited target data. Fig. 4(c) and (d) present the corresponding intra-class distances. We can observe that the TDA(O) maintains the smallest distances than all baseline methods.

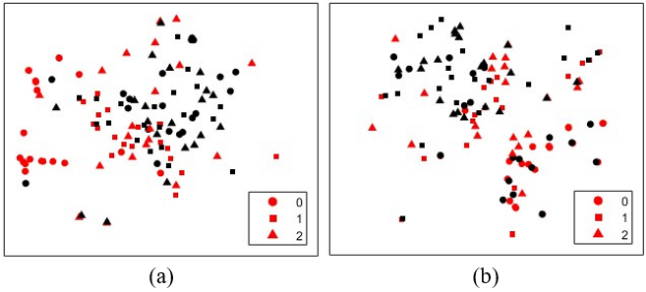

**Fig. 2.** Class-level t-SNE plots of MNIST-M data. Red and black points represent the source MNIST, and the target MNIST-M, respectively using (a) TAISL, and (b) TDA(O).

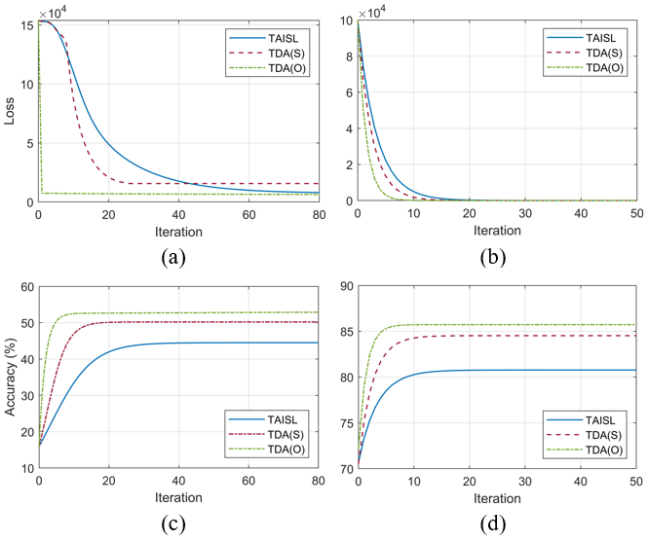

**Fig. 3.** Training loss for (a) image data and (b) audio data, and classification accuracy for (c) image data and (d) audio data.

To visualize the impact of the oblique constraint, we present the matrices $M_k M_k^T$ obtained from TAISL, TDA(S) and TDA(O) in Fig 5. While the matrices from Stiefel (TAISL, TDA(S)) display a diagonal structure, those from oblique (TDA(O)) contain non-negligible off-diagonal elements. This indicates that the proposed oblique constraint enhances interdependence among projection directions, leading to improved capturing of domain-specific variations.

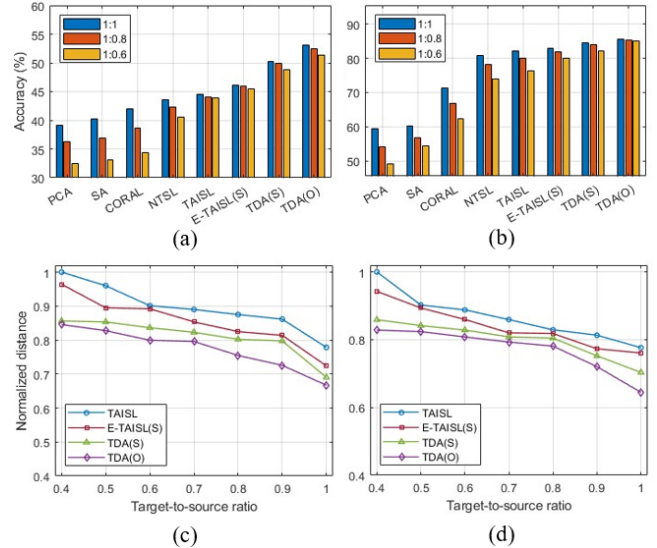

**Fig. 4.** Classification accuracy for (a) image data and (b) audio data, and intra-class distance for (c) image data and (d) audio data, across varying target-to-source sample ratios.

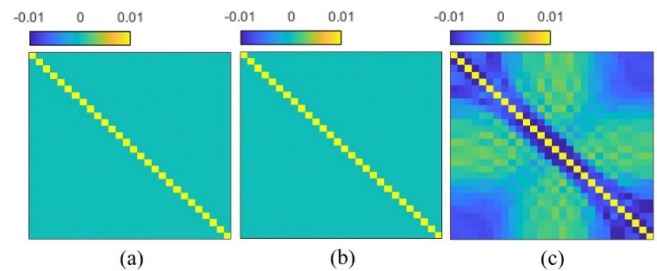

**Fig. 5.** Visualization of $M_2 M_2^T$ obtained from (a) TAISL, (b) TDA(S) and (c) TDA(O).

## 4. CONCLUSIONS

In this work, we introduced a novel, unified tensor-based domain adaptation framework that effectively aligns source and target tensors within an invariant subspace via iterative optimization of alignment matrices. By imposing constraints on the oblique manifold, our approach provides greater flexibility and adaptability compared to traditional methods. The incorporation of variance-preserving regularization further enhances robustness. Extensive experiments validated that our method significantly accelerates DA processes and achieves superior classification accuracy, outperforming existing state-of-the-art techniques. These results demonstrate the efficacy and versatility of our framework, promising improved performance in complex domain adaptation scenarios.